\documentclass[conference]{IEEEtran}
\IEEEoverridecommandlockouts
\usepackage{cite}
\usepackage{amsmath,amssymb,amsfonts}
\usepackage{algorithmic}
\usepackage{graphicx}
\usepackage{multirow}
\usepackage{textcomp}
\usepackage{xcolor}
\usepackage{float}
\usepackage{caption}
\usepackage{subcaption}
\def\BibTeX{{\rm B\kern-.05em{\sc i\kern-.025em b}\kern-.08em
    T\kern-.1667em\lower.7ex\hbox{E}\kern-.125emX}}
\begin{document}

\title{Feature Selection for Multivariate Time Series via Network Pruning}

\author{\IEEEauthorblockN{Kang Gu}
\IEEEauthorblockA{\textit{Dartmouth College} \\
kang.gu.gr@dartmouth.edu}
\and
\IEEEauthorblockN{Soroush Vosoughi}
\IEEEauthorblockA{\textit{Dartmouth College} \\
soroush.vosoughi@dartmouth.edu}
\and
\IEEEauthorblockN{Temiloluwa Prioleau}
\IEEEauthorblockA{\textit{Dartmouth College} \\
temiloluwa.o.prioleau@dartmouth.edu}
}

\maketitle

\begin{abstract}
In recent years, there has been an ever increasing amount of multivariate time series (MTS) data in various domains, typically generated by a large family of sensors such as wearable devices. This has led to the development of novel learning methods on MTS data, with deep learning models dominating the most recent advancements. Prior literature has primarily focused on designing new network architectures for modeling temporal dependencies within MTS. However, a less studied challenge is associated with high dimensionality of MTS data. In this paper, we propose a novel neural component, namely Neural Feature Selector (NFS), as an end-2-end solution for feature selection in MTS data. Specifically, NFS is based on decomposed convolution design and includes two modules: firstly each feature stream\footnote{A stream corresponds to an univariate series of MTS} within MTS is processed by a temporal CNN independently; then an aggregating CNN combines the processed streams to produce input for other downstream networks. We evaluated the proposed NFS model on four real-world MTS datasets and found that it achieves comparable results with state-of-the-art methods while providing the benefit of feature selection. Our paper also highlights the robustness and effectiveness of feature selection with NFS compared to using recent autoencoder-based methods.
\end{abstract}


\section{Introduction}
Multivariate time series (MTS) data has been drawing increasing attention from researchers in various domains, including industrial design \cite{Industrial-Design}, retail \cite{TADA} and medicine \cite{MTS-with-Missing-values}. However, two major challenges associated with modelling MTS data are: i) the time span can be up to several years, which makes long-term dependencies hard to capture (e.g. stock market \cite{stock_market}), ii) high dimensionality in many MTS datasets (e.g. mobile sensing \cite{OhioT1DM}) can cause over-fitting and poor understanding of underlying feature correlations \cite{Feature-selection-review}.  Majority of recent work \cite{time-series-review} on MTS has focused on developing novel deep learning models to address the first challenge. However, efforts to address the challenge of high dimensionality is still limited.
Effective feature selection (e.g. remote sensing \cite{FS_satellite} and traffic \cite{FS_traffic}) can boost the final performance of deep learning models, as well as reduce the risk of overfitting and the cost of computation.


To resolve the challenge of high dimensionality in MTS data, feature extraction\cite{autoencoder1} and feature selection methods \cite{filter1,wrapper1,embedding1} are imperative. However, feature construction methods that create new features for use does not satisfy the requirement of interpretability \cite{agnostic-feature-selection}, thus making feature selection critical in developing interpretable machine learning solutions. 
A detailed review of feature selection methods is represented in the paper by Saeys et al. \cite{Feature-selection-review}. Unlike with static variables, achieving feature selection with MTS data includes handling larger volumes of data with dependencies across variables. 


In this paper, we introduce - Neural Feature Selector (NFS) - as an end-2-end solution for feature selection in MTS data. 
NFS employs a decomposed convolution design, which is includes a temporal CNN module and an aggregating CNN module. Specifically, each stream within MTS is first processed independently by the temporal CNN module, then an aggregating CNN will extract channel-wise information from the processed streams.
Unlike existing feature selection methods such as traditional approaches\cite{CleVer,Mutual-Information,Causual-Discovery} and recent attention-based approaches \cite{DA-RNN,temporal-fusion-transformer}, we formulate the proposed approach as a network pruning problem.
 Given that each feature stream corresponds to an individual path connected to the aggregating CNN, an importance score for each path will be learned with gradient descent. The importance scores are then ranked at the end of model training to eliminate redundant connections. 
Key contributions of this paper are:
\begin{enumerate}
    \item We propose NFS, a novel feature selection framework for MTS. NFS can be trained jointly with various downstream models (e.g. Transformer) to perform feature selection in one training session. Therefore, NFS requires minimal overhead.

  \item Experimental results on four publicly-available datasets (OhioT1DM \cite{OhioT1DM}, Favorita \cite{TADA}, PhysioNet 2012 \cite{MTS-with-Missing-values} and Face Detection \cite{Unsupervised_TSL} ) show that NFS can boost the performance of state-of-the-art MTS models including Resnet, MCNN, MCDCNN, T-leNet and Transformer by joint training and selecting discriminative features.
 Noticeably, combination of NFS and MCDCNN achieved an accuracy of 0.636 on the Face Detection dataset, exceeding the most recent work by more than 10\%.
  
    
    \item We formulate feature selection as a network pruning problem, which provides a new perspective of conducting feature selection for deep learning methods.
    
    \item We make the code and the experimental data for this paper available for general use. (Submitted as supplementary for review)
  
\end{enumerate}
\section{Related Work}

\subsection{Deep Learning for MTS Analysis}
With the increasing volume of temporal data, various DL-based MTS algorithms have been proposed recently \cite{time-series-review}. For example, ResNet, which achieved huge success in Computer Vision tasks, has been adapted for MTS classification \cite{ResNet}. It is still regarded as the state-of-the-art model in the time series domain. MCNN exploited features at different scales and frequencies by performing Window Slicing technique on input data, which consisted of three transformations, namely identity mapping, down-sampling, and smoothing \cite{MCNN}.  Moreover, NPE \cite{NPE} and MCDCNN \cite{MCDCNN} are more relevant to our approach because they both employed a decomposed convolution design for MTS data, with the convolutions applied independently (in parallel) on each dimension (or channel) of the input MTS. Such design can capture both temporal and channel-wise information effectively. 
While NFS is based on the design of NPE, we explore its potential of feature selection rather than pattern extraction. More specifically, NFS is designed to be a universal feature selection framework for MTS, while the above-mentioned models are only focused on classification/regression.

Besides the above supervised MTS classification/regression methods, unsupervised MTS representation learning has also been explored \cite{Unsupervised_TSL}. Franceschi \textit{et al.} proposed a novel triplet loss inspired by word2vec, which could produce high-quality MTS embeddings for classification. Extensive experiments showed that their method achieved comparable results with state-of-the-art models.

\subsection{Feature Selection for MTS Analysis}
In general, traditional feature selection algorithms are categorized into: the filter approach \cite{filter1}, the wrapper approach \cite{wrapper1} and the embedding approach \cite{embedding1}. Most of these methods are not compatible with deep learning framework. In fact, feature selection methods for deep neural networks are still little studied. Qin \textit{et al.} proposed a dual-stage attention-based RNN model (DA-RNN) \cite{DA-RNN}, for time series forecasting. It relied on the first-stage and second-stage attention mechanisms to select relevant input features and hidden states respectively. Likewise, Li \textit{et al.} \cite{Logsparse_Trm} proposed convolutional self-attention which can identify important local context to enhance Transformer's performance on MTS data. Their model achieved lower training loss and better forcasting accuracy compared with RNN-based models. Moreover, Doquet \textit{et al.} \cite{agnostic-feature-selection} combined an AutoEncoder with structural regularizations to achieve the aim of
feature selection. But this method can not guarantee the performance of selected features in downstream tasks due to the unsupervised setting. Compared to the aforementioned work, the proposed NFS approach has two advantages: i) unlike attention mechanisms, NFS is a flexible head component that can be plugged into various downstream models, which requires no modifications of internal architectures, and ii) NFS treats feature selection as a network pruning problem which is a novel approach for performing feature selection.
\section{Neural Feature Selector}

In this section, we describe the architecture of our proposed NFS method and explain how it can be used for feature selection.

\begin{figure*}
    \centering
    \includegraphics[width=.9\textwidth]{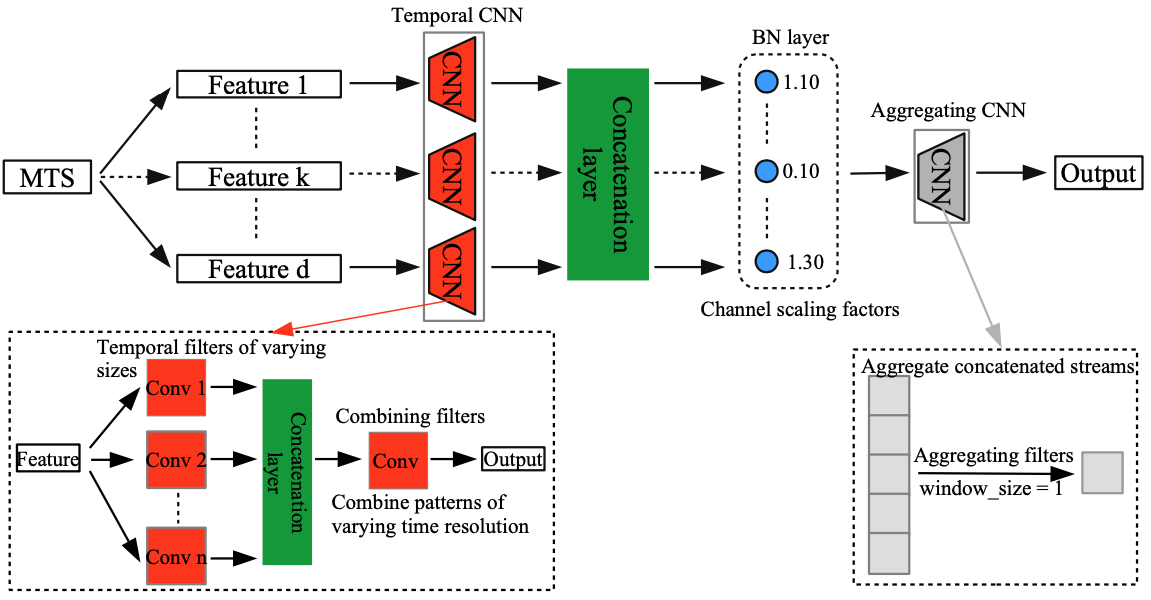}
    \caption{An illustration of Neural Feature Selector. The internal structures of Temporal CNN and Aggregating CNN are revealed for better understanding. We use dotted line to represent the path which will be discarded after evaluating the importance of each feature.} 
    \label{fig:NFS}
\end{figure*}

\subsection{Architecture}

Fig. \ref{fig:NFS} presents an overview of the architecture of our proposed NFS module. The input MTS data is first split into separate streams. Then each stream of data is processed independently by a temporal CNN tailored to extract unique temporal dependencies in corresponding feature streams. 
Next, the processed streams are concatenated and passed to a batch normalization layer, followed by an aggregating CNN. Finally, the output can be utilized by any downstream network (e.g. LSTM) to conduct a specific task such as classification/ regression. A formal description is as below:

Given that $X_{1:T}$ is the $d$ dimensional input temporal sequence, the output of each temporal CNN $x^{i}_{1:T}$ is $k$ dimensional, where $k$ is the number of filters in a temporal CNN module. The concatenation layer then stacks the $d$ streams up from the preceding part to obtain $\hat{X}_{1:T} \in R^{T\times kd}$. Then Batch normalization (BN) layer is applied to normalize the stacked streams. Finally, the aggregating CNN employs $n$ filters of size $1$ to map  the $k*d $ dimensional features  to dimensionality of $n$, which is compatible with the downstream network.

\subsection{NFS for Feature Selection}

\begin{figure*}
    \centering
    \includegraphics[width=.9\textwidth]{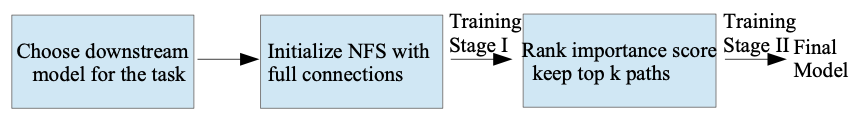}
    \caption{Feature selection schema with our Neural Feature Selector.} 
    \label{fig:schema}
\end{figure*}

There are two beneficial attributes of the proposed NFS brought on by the decomposed design. Firstly, each feature forms an individual path on the NFS graph hence dropping out unique features do not affect the connectivity of other paths. Secondly, the dimensionality of the output only depends on the number of aggregating filters. This means reduction in the number of features does not change the input size of downstream network. This enables feature selection with minimal overhead. More specifically, we treat feature selection as a part of the network training process. Fig. 2 \ref{fig:schema} shows a feature selection schema of our proposed NFS module. For a given MTS task, model training (e.g. NFS combined with recurrent neural network) always starts with the whole feature set, then we can examine the importance scores of features to keep the $k$ most informative paths in the NFS graph. Finally, a more compact model is re-trained with the selected features to obtain comparable or better results.

\section{Methodology}


In this section, we first formulate the problem mathematically, then proceed to explain the calculation of feature importance score. The implementation details of NFS will be elaborated on at the end of this section.

\subsection{Problem Formulation}
The objective of training NFS is to identify important feature subsets from all the candidates. For an arbitrary MTS task, the input is defined as a $d-$dimensional sequences: $X_{1:T} = [X_1, X_2, ..., X_T]$, where $T$ is the length of time series. At each time step, $X_i = [x_i^1 \oplus x_i^2, ..., \oplus x_i^d]$, $i \in[1,T]$, in which $\oplus$ is the symbol of concatenation. To perform feature selection, an importance score $\alpha_j$ is associated with corresponding feature stream $[x_1^j, x_2^j, ..., x_T^j]$ and is updated during training. The streams (features) with top $k$ importance scores will be chosen as final feature subset. 
$$\hat{X}_{1:T} =  [\hat{X_{1}}, \hat{X_{2}}, ..., \hat{X_{T}} ]   $$
$$\hat{X_i} = [x_i^{arg1}, x_i^{arg2}, ..., x_i^{argk}]$$
$$[arg1, arg2, ..., argk] = argmax([\alpha_1, \alpha_2, ..., \alpha_d], k)$$

where we use $argmax([\cdot], k)$ to return the indices of the top $k$ scores. 

\subsection{Importance Score}
To avoid introducing additional complexity to our NSF architecture, we choose to reuse the Batch Normalization (BN) layer as importance score layer \cite{networkslimming}. BN \cite{BN} has been adopted widely by CNN to achieve fast convergence and better generalization performance. let $z_{in}$ and $z_{out}$ be the input and output of BN layer. More specifically, we use $z_{in}^{i}$ and $z_{out}^{i}$ to represent the $i_{th}$ stream (feature), $i \in [1, d]$ :
$$z_{norm}^i = \frac{z_{in}^i - {\mu}_i}{\sqrt{({\sigma}_i)^2 + \epsilon}}; z_{out}^i = \alpha_{i} z_{norm}^i + \beta_i $$

Where $\mu_i$ and $\sigma_i$ are the mean and standard derivation values of $i_{th}$ stream over the batch, $\alpha_i$ and $\beta_i$ are trainable affine parameters. Moreover, $\alpha_i$ is also reused as the importance score for $i_{th}$ feature. As in Fig \ref{fig:NFS}, the channel scaling factors are the values of the importance scores $[\alpha_1, \alpha_2, ..., \alpha_d]$.

\subsection{Training with Importance Score Penalty}
We apply Lasso regularization \cite{Lasso} to impose sparsity on these importance scores. The idea is to make only a few streams have high scores, while push other scores to be close to zero. The training objective of our approach is as below:
$$ L = \sum_{x,y} l( f(x,W),y) + \gamma\sum_{i=1}^{d} |\alpha_i| $$

Where $x$ and $y$ denote the train input and the target, $W$ denotes the trainable weights of the network.
For regression task, the loss function $l(\cdot)$ can be Mean Squared Error. While for classification task, it can be Cross Entropy. The last term is the Lasso penalty, which is also known as L1-regularization. $\gamma$ is set to be $0.001$ through all experiments.

\section{Experimental Settings}
In this section, we will begin with describing the datasets and the corresponding preprocessing procedures. The overall experimental setup will be presented at the end.  

\begin{table}
\centering
\caption{Summary of the benchmarks in our experiments}\label{benchmarks}
\resizebox{.48\textwidth}{!}{
\begin{tabular}{|l|l|l|l|l|}
\hline
Dataset &  Time Range & Granularity &  Variables &  Type  \\
\hline
OhioT1DM \cite{OhioT1DM} &  8 weeks &  5 minutes  & 19  & mobile sensing \\
\hline
Favorita &  365 days &  1 day  & 15  &  commercial  \\
\hline
PhysioNet 2012 & 48 hours&  1 hour  & 37  & medical \\
\hline
Face Detection & 1.5 seconds  & unknown   &  144  & medical   \\
\hline
\end{tabular}}
\end{table}
\begin{table}
    
    \centering
    \caption{Hyperparameters for Training}
    \label{table_hyperparameters}
    \centering
    \resizebox{.48\textwidth}{!}{
    \begin{tabular}{|c|c|c|c|c|}
        \hline
         Parameter Name & OhioT1DM & Favorita & PhysioNet  & FaceDetection\\
        \hline
        \# of selected features &5/19  &5/15 & 16/37 & 32/144\\
        \hline
         sequence length &   12  & 365  & 48 & 62\\
        \hline
         coefficient of $L_{2}$-penalty & 0.01  &0.01 & 0.01 & 0.01\\
        \hline
         coefficient of architecture penalty& 0.001 &  0.001  & 0.001 & 0.001\\

        \hline
         train epoch  & 120 & 100 & 100 & 100\\
       \hline
       batch size  & 256 & 128  & 128 & 128 \\
        \hline
        Evaluation Metrics &RMSE  & MAE & AUC & Accuracy\\
       \hline        
    \end{tabular}
    }
\end{table}
\subsection{Datasets}
To validate the general learning capability of NSF, we consider four public MTS datasets, namely \textbf{OhioT1DM}, \textbf{Favorita}, \textbf{PhysioNet 2012} and \textbf{Face Detection}, from varying domains. The statistics of aforementioned datasets are provided in Table \ref{benchmarks}. Their properties are introduced below:

\noindent \textbf{OhioT1DM} A publicly available dataset \cite{OhioT1DM} which consists of eight weeks of CGM, insulin, fitness tracker, and self-reported life-events data from 6 subjects with type 1 diabetes (ages: 40 to 60 yrs.). Self-reported life-event data (e.g. exercise) was collected via a smartphone app and physiological data (e.g. heart rate) was collected with a Basis fitness band. This dataset was already split into a train and test set which was used in this work to facilitate fair comparison with related research papers. We follow the steps in \cite{DRNN} for preprocessing. 

There are a lot of missing CGM values in both the training and test set of OhioT1DM. Informed by prior work \cite{DRNN}, we performed a first-order interpolation and first-order extrapolation on training and test sets, respectively. A median filter was then employed to remove false spikes after the interpolation process. With this dataset, the input to models described in this work consists of 19 dimensions, including  glucose\_level, basal, bolus, self-reported carbohydrate estimates, and other related life-event data which are described in the original paper \cite{OhioT1DM}. \\[12pt]

\noindent \textbf{Favorita} A real-world sales dataset provided by a large Ecuadorian-based grocery retailer\footnote{https://www.kaggle.com/c/favorita-grocery-sales-forecasting/data}. The dataset contains daily sales volume of all products from 56 grocery stores located in different cities.
Additionally, the product/store information as well as related events (e.g. national holiday and oil price) are provided as supplementary information for sales prediction. The original Favorita dataset covers the time from 1 January 2013 to 15 August 2017, but we
only use the portion from 15 August 2016 to 15 August 2017 as in \cite{TADA}, where more details of experimental settings can be found.

Each time series is essentially a log file for a specific item in Favorita. As in \cite{TADA}, we don't split different products up for training and testing, because it is impossible to make prediction for unseen products in realistic business. Thus, we first randomly take 20\% of total samples for validation. Then for the rest samples, we apply a "walk-forward" strategy to split the data. Suppose the prediction horizon is $\Delta$ days ($\Delta$ is set to be $[2, 4, 8]$ respectively). During train stage, we use the range of [1, 365 -2 $\Delta$]  to predict the sales of [$366 - 2 \Delta$, $365 - \Delta$]. For evaluation, We then use the range of [$\Delta +1$, $365 - \Delta$] to produce the sales prediction of [$366 - \Delta$, 366] to test the accuracy. \\[12 pt]

\noindent \textbf{PhysioNet 2012} This dataset is from PhysioNet Challenge 2012\footnote{https://physionet.org/content/challenge-2012/1.0.0/}, which is composed of 8000 intense care unit (ICU) records from different hospitals. Each record is a MTS of roughly 48 hours and contains 37 variables that reflects the patient's physiological sate. We use the Training Set A subset (contains 4000 patients) to conduct the task of morality prediction as in \cite{MTS-with-Missing-values} 

We randomly split the data into  training set with 3200 samples and test set with 800 samples. However, the data distribution is severely uneven. There are 3485 patients with the label 0, while only 515 patients are labeled as 1. We apply a up-sampling algorithm, namely ADASYN \cite{ADASYN} to redress the imbalance. Finally, The training set ends up with 2786 negative samples and 1414 positive samples. \\[12pt]

\noindent \textbf{Face Detection}
This dataset is from the UEA MTS classification archive\footnote{http://www.timeseriesclassification.com/dataset.php}. The reason of including it in our experiment is twofold: 1) Its dimensionality is relatively high, which poses challenge to feature selection. 2) There are enough train/test samples in this dataset, so that the results are not influenced by overfitting. This dataset consists of MEG signals
collected from 16 subjects (10 for train and 6 for test). The task is to determine whether the subject has been shown the picture of a face or a scrambled image based on MEG. 
There are about 580-590 trials per patient, yielding 5890 train trials and 3524 test subjects. Each trial consists of 1.5 seconds of MEG recording.
\subsection{Experimental Setup}
The training hyperparameters can be found in Table \ref{table_hyperparameters}. All the experiments were conducted on a machine with 2.6 GHz Intel Core i7, 16 GB RAM and Nvidia Geoforce RTX 2060. Model training can typically be finished within a hour. For more implementation details, please refer to the code submitted as supplementary. 
\\
\\
\noindent \textbf{Downstream Models} Since NFS is a head component, we first consider LSTM, the most prevalent model of time seris , as well as four more recent CNN-based models, including ResNet \cite{ResNet}, MCNN \cite{MCNN}, MCDCNN \cite{MCDCNN} and t-LeNet \cite{time-series-review}, as first five downstream models. 
Moreover, we are also aware of the recent success of Transformer-based models \cite{temporal-fusion-transformer} in time series domain. Thus, we extend the scope of downstream models by the vanilla Transformer model from the original paper \cite{DBLP:journals/corr/VaswaniSPUJGKP17}. 

We represent the complete models by NFS+LSTM, NFS+ResNet, NFS+MCNN, NFS+ MCDCNN, NFS+t-LeNet and NFS+Transformer respectively. The parameters for each model can be found in the Appendix.



\begin{table*}[h]
    \caption{Results on the OhioT1DM \& Favorita. On OhioT1DM, \textbf{5/19} features are selected. On Favorita, \textbf{5/15} features are selected}
    \label{tab:regression_result}
    \centering

    \begin{tabular}{|l|c|c||c|c|c|c|}
    
    \hline
      \parbox{27mm}{\multirow{3}{*}{ Models= } } &  
    & OhioT1DM   &  \multicolumn{3}{c|}{Favorita} \\
    \cline{3-6}
     & & RMSE (\textit{mg/dL})  & \multicolumn{3}{c|}{MAE (\textit{unit sales}) }\\
    \cline{3-6}
    & &   $\Delta = 30$ mins  &   $\Delta = 2 $ days &  $\Delta = 4$& $\Delta = 8$  \\
    \hline    
    
   Midroni \textit{et al.} \cite{XGBoost} & XGBoost & 19.32  & - & - & - \\
   Gu \textit{et al.} \cite{NPE} & NPE+LSTM & 17.80 & - & - & -\\
   \hline
   Lai \textit{et al.} \cite{LSTnet} & LSTNet & - &7.419 & 7.982& 8.729   \\   
   Chen \textit{et al.} \cite{TADA} & TADA & - & 6.955 & 7.323 & 7.422 \\
    \hline
    
    \parbox{20mm}{\multirow{7}{*}{  Baselines } } 
  
    &ResNet   & 28.41 &8.272 & 8.713  & 9.687\\
    &MCNN    &28.68  & 5.921  & 6.252 & 6.335\\
    &MCDCNN   & 28.17  &\textbf{5.644}  & \textbf{5.903} & \textbf{6.248}\\
    &t-LeNet  & not converge  &7.243   & 7.532 & 7.677 \\
    & Transformer & 21.31 &  7.453   &  7.900    & 7.945 \\
    \hline
    
    \cline{2-6}
    \parbox{20mm}{\multirow{3}{*}{  NFS+LSTM  } } 
    & Full feature set  & 17.80 & 4.877  & 4.900  &  5.112\\
     &  Selected by AgnoS &  19.52   & 5.044 & 5.066     & 5.279 \\
    & Selected by NFS & $\mathbf{17.50^\dagger}$  & $\mathbf{4.441^\dagger}$ & $\mathbf{4.584^\dagger}$  &
    $\mathbf{4.791^\dagger}$ \\
    \cline{2-6}
    \parbox{20mm}{\multirow{3}{*}{ NFS+ResNet  } } 
     & Full feature set & 28.30   & 7.714 & 8.810 & 9.269 \\
    
    &  Selected by AgnoS & \textbf{24.16}    & \textbf{5.289} & \textbf{6.158} & \textbf{7.293} \\
    & Selected by NFS  &  24.38 &6.413  &7.443 & 7.518  \\
    \cline{2-6}
    \parbox{20mm}{\multirow{3}{*}{  NFS+MCNN  } } 
     & Full feature set & 26.40   & 5.985 &6.037   & 6.206 \\
    
    &  Selected by AgnoS &  \textbf{22.50}  &  5.551& 5.631 & 5.776  \\
    & Selected by NFS  & 22.67  & \textbf{5.242} & \textbf{5.272}  & \textbf{5.429} \\
    \cline{2-6}
     \parbox{20mm}{\multirow{3}{*}{  NFS+MCDCNN  } } 
     & Full feature set &   25.02 &4.936 & 5.036  & 5.120  \\
    
    &  Selected by AgnoS & 21.45   &5.292  & 5.326& 5.423  \\
    & Selected by NFS  &  \textbf{20.82}& \textbf{4.827} & \textbf{4.955}  & \textbf{5.002} \\
    \cline{2-6}
     \parbox{20mm}{\multirow{3}{*}{  NFS+t-LeNet  } } 
     & Full feature set & 28.05   &7.027 &  7.494  & 7.892 \\
    &  Selected by AgnoS & 24.77  &6.160 & 6.281  & 6.466   \\
    & Selected by NFS & \textbf{22.33} & \textbf{5.184} & \textbf{5.240}  & \textbf{5.267} \\
    \cline{2-6}
    \parbox{20mm}{\multirow{3}{*}{  NFS+Transformer  } } 
     & Full feature set & 19.92    & \textbf{5.147} & \textbf{5.108} & \textbf{5.330}  \\
    &  Selected by AgnoS & 39.43 & 8.058 &  8.070  & 8.542   \\
    & Selected by NFS & \textbf{18.31}  & 5.311 & 5.408 & 5.497  \\
    \cline{2-6}
    \hline

    \hline    
    \end{tabular}
    
\end{table*}

\section{Results}
The goal of our experiments is twofold: i) We demonstrate that the NFS framework is flexible enough to be adapted to various downstream models (ii) To evaluate the effectiveness of NFS, we display the results before and after feature selection for comparison. Additionally, the results of AgnoS \cite{agnostic-feature-selection}, a recent AutoEncoder-based feature selection method, are also provided. Notably, we train every model on each task $5$ times to obtain average performance. 

Note that "Full feature set", "Selected by AgnoS" and "Selected by NFS" in Table \ref{tab:regression_result} \& \ref{tab:classification_result} refer to using NFS module with the corresponding feature subsets.

\subsection{Results on OhiT1DM \& Favorita}
For simplification, we use $\Delta$ to represent prediction horizon. 
As shown in Table \ref{tab:regression_result}, NFS boosted the performances of all downstream models on both datasets. Specifically, NFS+t-LeNet model achieved a RMSE of 28.05 \textit{mg/dL} with full feature set on OhioT1DM dataset, while only using feature subset selected by NFS reduced the RMSE by 5.72. Besides, NFS+LSTM model outperformed the state-of-the-art NPE+LSTM by 0.3 after NFS performed feature selection. As for Favorita dataset, NFS brought down the MAE of t-LeNet by more than 2 via feature selection. More noticeably, the MAE of NFS+RNN model was even reduced by about 4 after feature selection of NFS.


Comparing AgnoS with our NFS, experimental results showed that NFS outperformed AgnoS on four out of six downstream models on both OhioT1DM and Favorita. Even in cases where AgnoS won, the gaps between them were narrow. Noticeably, NFS outperformed AgnoS by more than 20 of RMSE on OhioT1DM with Transformer as downstream model.
On average, NFS beat AgnoS by 0.86 of RMSE and 0.60 of MAE on OhioT1DM and Favorita respectively. Although both our NFS and Agnos can filter redundant features, NFS constantly outperformed the results of full feature set on both datasets. While AgnoS failed to maintain the accuracy on two out of seven models on both datasets. Since AgnoS is an unsupervised feature selection method, it's reasonable that the resulting feature subset may not be the optimal for the specific task. While our NSF produces final feature subset based on gradient information during training, the results are constantly promising.

\begin{table}[h]
    \caption{Results on PhysioNet 2012 \& Face Detection. On PhysioNet 2012, \textbf{16/37} features are selected. On Face Detection, \textbf{32/144} features are selected.}
    \label{tab:classification_result}
    \centering
    \resizebox{.48\textwidth}{!}{
    \begin{tabular}{|l|c|c||c|}
    
    \hline
    \parbox{25mm}{\multirow{2}{*}{models}}
    &    &     PhysioNet & Face Detection \\
    \cline{3-4}
      &  & AUC Score & Accuracy Score  \\

    \hline    
    
    \parbox{25mm}{\multirow{3}{*}{ Che \textit{et al.} \cite{MTS-with-Missing-values} } } 
    & GRU & 0.8226  & -\\
    & GRU (PCA reduced) & 0.8116  & -\\   
    & GRU-D & 0.8424  & -\\
    \hline
    Franceschi \textit{et al.} \cite{Unsupervised_TSL} & Encoder & - &0.528 \\
   Fang \textit{et al.} \cite{SNN}  & SNN  & - &0.57 \\
    \hline
    
    \parbox{25mm}{\multirow{7}{*}{  Baselines } } 

    & ResNet & 0.8595 &  0.541\\
    & MCNN &  0.8805  &  0.500\\
    & MCDCNN & 0.8720 &  \textbf{0.626}\\
    & t-LeNet & \textbf{0.9016} & 0.588 \\
    & Transformer  & 0.8755  & 0.562 \\
    \hline

    \cline{2-4}

    \cline{2-4}
    \parbox{25mm}{\multirow{3}{*}{  NFS+LSTM } } 
    &Full feature set  & 0.8731 & 0.580 \\
    &Selected by AgnoS &  0.8819 &  0.514\\
    & Selected by NFS & \textbf{0.9117} & \textbf{0.588}  \\
    \cline{2-4}
    \parbox{25mm}{\multirow{3}{*}{NFS+ResNet } } 
    &Full feature set  &  0.8713 & 0.527  \\
    &Selected by AgnoS &  0.8716 & 0.511 \\
    & Selected by NFS & \textbf{0.8810} & \textbf{0.545}  \\
    \cline{2-4}
    \parbox{25mm}{\multirow{3}{*}{NFS+MCNN } } 
    &Full feature set  & 0.8814 &  0.5\\
    &Selected by AgnoS & 0.8919 & 0.5\\
    & Selected by NFS & \textbf{0.8925} & 0.5\\
    \cline{2-4}
    \parbox{25mm}{\multirow{3}{*}{NFS+MCDCNN   } } 
    &Full feature set  & 0.8801 & 0.629 \\
    &Selected by AgnoS & 0.8880 & 0.598   \\
    & Selected by NFS & \textbf{0.9026} & $\mathbf{0.636^\dagger}$ \\
    \cline{2-4}
    \parbox{25mm}{\multirow{3}{*}{NFS+t-LeNet } } 
    &Full feature set  & 0.9045 & 0.607\\
    &Selected by AgnoS & 0.9032 & 0.597\\
    & Selected by NFS & $\mathbf{0.9179^\dagger}$ & \textbf{0.610}
    \\ 
    \cline{2-4}
        \parbox{25mm}{\multirow{3}{*}{  NFS+Transformer } }
 &Full feature set  & 0.8942 &  0.516\\
  &Selected by AgnoS & 0.8900  &  0.605\\
    & Selected by NFS & \textbf{0.9114}  & \textbf{0.621}  \\
    \hline

    \hline    
    \end{tabular}
    }
\end{table}

\subsection{Results on PhysioNet 2012 \& Face Detection}
Table \ref{tab:classification_result} summarizes the results obtained on PhysioNet 2012 and Face Detection. NFS still boosted the performance of downstream models remarkably on both datasets. For PhysioNet 2012, NFS improved the AUC score by 0.06 for the RNN downstream model via feature selection. Besides, NFS+t-LeNet achieved the highest AUC of 0.9179 using feature subset selected by NFS, exceeding the recent work by about 0.08. As for Face Detection, the highest accuracy of 0.636 was obtained by NFS+MCDCNN model trained with NFS feature subset. Compared with the recent SNN, our model improved the performance by more than 10\%. Meanwhile, NFS+Transformer achieved comparable results against the best models on both datasets.
Note that MCNN didn't converge on Face Detection dataset, thus it should be ignored.  

Regarding the impacts of feature selection, we found that NFS achieved the best results on all the seven downstream models on both datasets, thus beating AgnoS in all the cases. On average, NFS achieved an AUC score of 0.8860 on PhysioNet 2012, outperforming the average of AgnoS by about 0.02. 
After ignoring MCNN on Face Detection dataset, the average Accuracy score of NFS is 0.595, exceeding that of AgnoS by 0.05. Besides, feature subset selected by AgnoS impaired the results on all six models, which suggested that AgnoS might not work well on high-dimensional MTS dataset (Face Detection is 144-dimensional). However, our NSF can still identify informative features and enhance the performances for downstream models regardless of high dimensionality.

\subsection{Hyperparameter Sensitivity}

\begin{figure}
\begin{subfigure}{.5\textwidth}
  \centering
  \includegraphics[width=.95\linewidth]{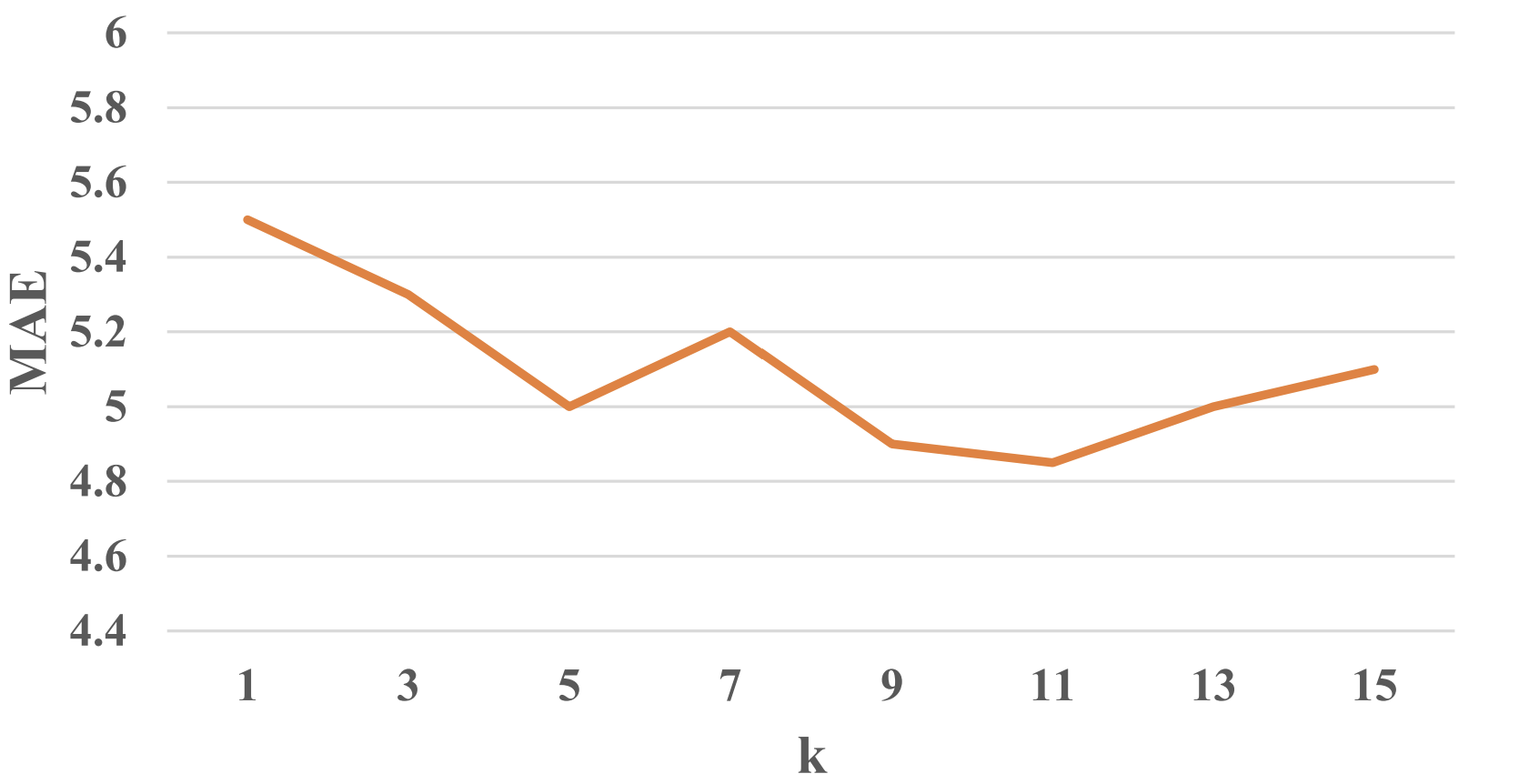}  
  \caption{Favorita}
  \label{fig:mcdcnn_favorita}
\end{subfigure}
\begin{subfigure}{.5\textwidth}
  \centering
  \includegraphics[width=.95\linewidth]{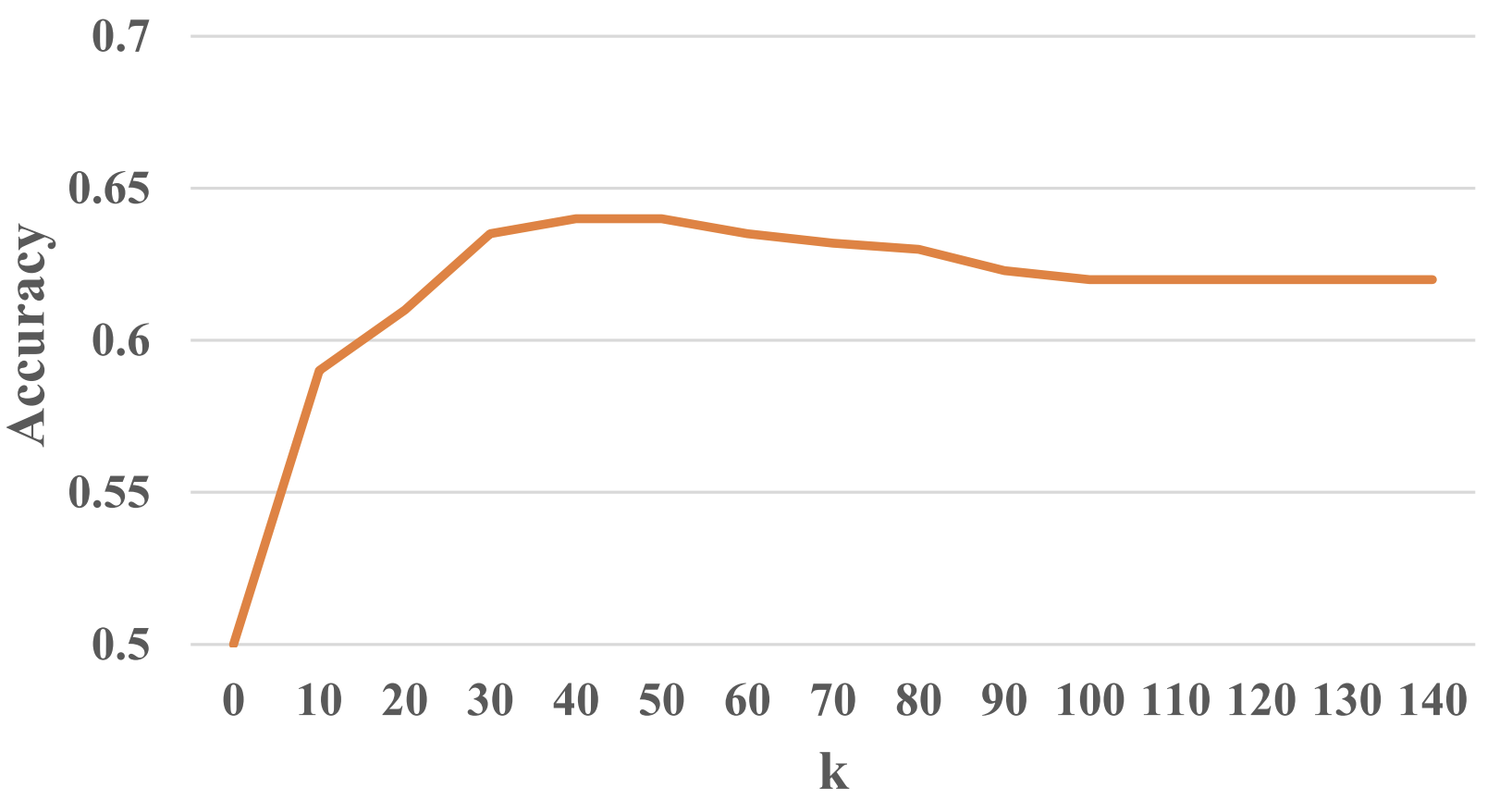}  
  \caption{Face Detection}
  \label{fig:mcdcnn_face}
\end{subfigure}
\caption{The sensitivity analysis of k (the number of selected features). We use NSF+MCDCNN as an example. (a) plots MAE, the lower value is better. (b) plots ACC, the higher value is better. }
\label{fig:sensitivity}
\end{figure}

Since k, the number of selected features, is the most important hyperparameter in our experiments, we have conducted sensitivity analysis as shown in Fig. \ref{fig:sensitivity}. According to Fig. \ref{fig:mcdcnn_favorita} \& \ref{fig:mcdcnn_face}, our model's performance first improves steadily with the increase of k, then gradually stabilizes when k is relatively large.

\section{Conclusion}

We propose NFS, an end-2-end feature selection framework for MTS data. NFS is a neural component based on decomposed convolution, in which temporal CNN processes each feature stream within MTS input independently and aggregating CNN combines all the streams to extract channel-wise information. NFS can also be viewed as a network pruning approach. When NFS is trained jointly with a downstream model, it learns to keep only the informative feature paths with gradient descent information and therefore it requires minimal overhead.

The evaluation on four real-world MTS datasets with varying dimensionality has shown the flexibility and effectiveness of NFS on various downstream networks. Besides, our NFS constantly outperformed AgnoS, an AutoEncoder-based feature selection algorithm, in terms of the performance of selected features.

\bibliographystyle{IEEEtran}
\bibliography{Reference}

\section{Appendix}

\subsection{Configuration of NFS}
The Configuration of NFS is shown in Table \ref{Configure_NFS}

\begin{table}[H]
\centering
\caption{Parameters of NFS }\label{Configure_NFS}
\begin{tabular}{|l|l|l|}
\hline
Module & Parameter  & Value \\
\hline
\parbox{25mm}{\multirow{3}{*}{ Temporal CNN } } 
    & size of filters in each branch  & $[2, 3, 4, 5]$  \\
    & \# of each filter  &   1 \\   
    &  padding &  same \\
    & stride & 1\\
    & dilation & 1 \\
    & activation & relu \\
    & regularizer & l2(0.01) \\ 
\hline
Concatenating Layer &  axis of concatenation  & 2    \\
\hline
\parbox{25mm}{\multirow{3}{*}{ Aggregating CNN } } 
    & size of filter &  1\\
    & \# of filter & 32 \\
    &  padding &  same \\
    & stride & 1\\
    & dilation & 1 \\
    & activation & relu \\
    & regularizer & l1(0.01) \\ 
\hline

\end{tabular}
\end{table}

\subsection{Configuration of CNNs}
The configuration of all the CNN models, including ResNet, MCNN, MCDCNN and t-LeNet, are listed in Table \ref{Configure_CNN}.

\begin{table}[H]
\centering
\caption{Parameters of CNNs }\label{Configure_CNN}
\begin{tabular}{|l|l|l|}
\hline
Model & Parameter  & Value \\
\hline
\parbox{25mm}{\multirow{3}{*}{ ResNet } } 
    & \# of blocks  &  3 \\
    & \# of conv layers in each block &  3\\
    & size of filters in each block & [8, 5, 3] \\
    & \# of filters in each block  &  [64, 128, 128] \\   
    &   skip connection in each block & Yes  \\
\hline
\parbox{25mm}{\multirow{3}{*}{ MCNN } } 
    & \# of conv layers  &  2 \\
    & \# of filters  &  [256, 256] \\   
    & size of filters  & [3, 3] \\
    &    pool factors   & [2, 2]  \\
\hline
\parbox{25mm}{\multirow{3}{*}{ MCDCNN } } 
    & \# of blocks  &  2 \\
    & \# of filters  &  [8, 8]  \\   
    & size of filters  &  [5, 5]\\
    &  pool factors   & [2, 2] \\
\hline
\parbox{25mm}{\multirow{3}{*}{ t-LeNet } } 
    & \# of conv layers  & 2 \\
    & \# of filters  &  [5, 20] \\   
    & size of filters  & [5, 5] \\
    &  pool factors      & [2, 4]  \\
\hline
\end{tabular}
\end{table}

\subsection{Configuration of LSTM}
\begin{itemize}
    \item \# of Layers : 1
    \item Hidden State: 32
    \item Regularizer: l1\_l2(0.01, 0.01)
    
\end{itemize}

\subsection{Configuration of Transformer}

\begin{table}[H]
\centering
\caption{Parameters of Transformer}\label{Configure_Transformer}
\begin{tabular}{|l|l|l|}
\hline
Module & Parameter  & Value \\
\hline
Embedded Layer & size of embedding & 128 \\
\hline
\parbox{40mm}{\multirow{3}{*}{ Multihead Attention Block } } 
    & size of hidden state  &  128 \\
    & \# of heads &   8 \\   
    & size of Q layer & 128 \\
    & size of K\_V layer & 256 \\ 
    & \# of blocks & 2 \\
\hline

\end{tabular}
\end{table}

\end{document}